\documentclass[letterpaper, 10 pt, conference]{ieeeconf}  % Comment this line out if you need a4paper

\IEEEoverridecommandlockouts                              % This command is only needed if 
\usepackage{graphicx}
\graphicspath{ {./images/} }
\usepackage{color,soul}
\usepackage{fancyhdr}

\usepackage[margin=1in,footskip=0.25in]{geometry}
\usepackage{url}

\usepackage{xcolor}
\usepackage{soul}

\title{\LARGE \bf Wind Gust Detection using Physical Sensors in Quadcopters
Capstone Project 2018
}

\author{Suwen Gu$^{1}$ and Menghao Lin$^{2}$% <-this % stops a space
\thanks{$^{1}$Suwen Gu is a Student of Computer Science Department, Data Analytics Program.
        Fordham University,
        {\tt\small sgu21@fordham.edu}}%
\thanks{$^{2}$Menghao Lin is a Student of Computer Science Department, Data Analytics Program.
        Fordham University,
        {\tt\small mlin48@fordham.edu}}%
}

\begin{document}

\maketitle
 \thispagestyle{fancy}
 \pagestyle{fancy}
 \renewcommand{\headrulewidth}{0pt}

%===============================================================================

\begin{abstract}
We propose the use of basic inertial measurement units (IMU) which contain sensors such as accelerometers and gyroscopes already on-board drones to detect the speed and direction of wind gusts. The ability to quickly sense wind gusts has many applications, the most notable of which is in flight assistance of the drone, where it may adjust motor power parameter to compensate for such external factors or steer the drone toward a safer direction. To illustrate the feasibility of the approach, we conducted studies to assess how reliably wind speed and wind direction can be detected while a quad-copter drone is hovering and then in-motion, using off-the-shelf classifiers. Empirical results with real-life data, collected on a micro aerial vehicle (MAV) in a physical room with a consumer-grade fan, show that (1.1) wind speed can be detected with high accuracy after training, not only on the same drone, but also across different drones of the same class, while (1.2) wind direction can be detected with high accuracy after training on the same drone, but with limited generalizability to other drones. (1.3) We demonstrate how real-time detection of wind speed, using offline trained models, is feasible and can be done with high accuracy. (2.1) Finally, we find the reason behind the lower accuracy for wind direction detection during the analysis of drones in-motion.
\end{abstract}

%===============================================================================

\section{Introduction}
\label{sec:intro}

Recently, unmanned aerial vehicles (UAVs), or drones, have increasingly been found in many real-life applications including search and rescue, warehousing, agricultural monitoring, and military-grade missions. In such applications, accurate navigation, control and keeping the drones safe while operating in dynamic and unknown environments remains one of the top priorities. Thus when wind gust arises in the flight path of the drone, we look to implement a cost effective and non-bloated method of detecting the speed and direction of the wind gust. Adding an anemometer will cost more and add weight, especially to Micro Air Vehicles (MAV, UAV) that are already weight and payload limited. In this work, we discuss and report the feasibility of using commodity physical sensors, e.g., accelerometers and gyroscopes, to help drones sense wind gusts. 

Accelerometers and gyroscopes have been used successfully in detecting human activities~\cite{ravi2005activity,weiss2016smartwatch}. In that line of work, such sensors are deployed in daily used devices such as smart phones~\cite{kwapisz2010cell}, smart watches~\cite{johnston2015smartwatch, weiss2016smartwatch}, or health devices, e.g., Fitbit~\cite{guo2013evaluation}. These works showed that as human users walk, eat, or jog, their body, hand, and arm movements will trigger such sensors to change their readings in manners that allowed highly accurate recognition of such activities.

Inspired by the success of such work, we propose to apply similar mechanisms to equip drones with the ability to detect wind gust. The intuition is that a key environment factor, e.g., air stillness, will most likely be disturbed by anything that comes with motion, such as wind. If the drone is able to accurately associate the existence of wind with the change in its sensory readings, it can track and infer whether there is any such entity around.

The organization of the paper is as follows. We first survey related works in detection for drones. Next detail different components of the environment sensing task, including it being cast as a classification problem and how it can be incorporated into a general wind-detection framework. We then discuss our empirical study in assessing how feasible this approach in real-life situations. The paper concludes with discussions on existing limitations of our study, as well as implications for future works.

%===============================================================================

\section{Related Works}
\label{sec:rel-works}

Environment sensing refers to the task of detecting other entities populating the same physical space in which the drone is operating. These entities can come with a physical body, such as walls, obstacles or other robots, or none at all, such as wind gusts, air turbulence, gas, or interfering radio-frequency signals.

Detecting visible entities have been the main subject of study in most research on environment sensing, the task of which is often achieved using vision-based approaches. For instance, Engel et al.~\cite{engel2012accurate} used a monocular SLAM system to detect objects in a scene, allowing the drone to navigate through unknown environments, without the need of artificial markers or external sensors, in areas where GPS cannot be accessed. Similarly, Ross et al.~\cite{ross2013learning} demonstrated how on-board monocular vision can be utilized to assist in the navigation and guidance of micro aerial drones through cluttered environments at low altitude, such as outdoor forests. Many of these achieve this through feature detection with SIFT/SURF(Speeded Up Robust Features) methods~\cite{mori2013first,aguilar2017obstacle}. In this line of work, applications of environment sensing often include localization of the drones~\cite{benini2013imu}, or for object detection and avoidance~\cite{mcguire2017efficient}. 

In contrast, our work aims to detect hazards that do not take a physical or visual form, such as wind gust, or even when obstacles are out of the field of view of the cameras on the drone. 

In detecting invisible entities such as gas leakages, a popular approach is the use of specialized sensors attached to their drones, such as micro-machined gas sensors for the purpose of monitoring and detection of gas leakage localization~\cite{rossi2014gas,rossi2016autonomous}. While these approaches work for special environment entities such as gas, it is unclear whether a sensor can be designed to reliably detect wind in real time especially for use on drones, as the propellers themselves may alter the airflow around the drone where the sensor would be located. Notably, most works in detecting wind gusts focus on the negation of such wind effects in controlling the drone, to create a stable flight~\cite{sydney2013dynamic,smeur2016gust}. Our work focuses more on detecting aspects of wind gusts that can be picked up by the drone's sensors, thus opening doors for other downstream tasks.

Environment sensing may be conducted on a broader scale than just a vicinity surrounding a drone. For instance, some works used UAVs equipped with cameras for remote sensing and monitoring of forestry and coastal environments~\cite{paneque2014small,klemas2015coastal}. Such approaches yield many important and practical applications, such as surveying forests, mapping canopy gaps and canopy height, and tracking forest wildfires \cite{tang2015drone}. 

Finally a paper from the Defense Science and Technology Group on Low Airspeed Measuring Devices for Helicopter~\cite{knight2003low}, though lacking in technical detail due to military related work. It did a great job in shading light on interesting methods in detecting low airspeed were mechanical (mechanical anemometer) and non-mechanical system (These include GPS and laser based systems, algorithm systems and neural networks).

%===============================================================================

\section{Physical Sensors in Drones}
\label{sec:physical-sensors}

In this work, we used CrazyFlie 2.0, a product by BitCraze \cite{BitcrazeAB2016}, for our experimentation. This MAV comes with the following on-board sensors:
\begin{itemize}
    \item 3 axis gyro (MPU-9250),
    \item 3 axis accelerometer (MPU-9250),
    \item 3 axis magnetometer (MPU-9250),
    \item Barometer (LPS25H).
\end{itemize}

To increase the range of sensors available, it is often possible to augment the CrazyFlie 2.0 drones with additional sensors. Some sample specialized decks, which can be mounted on a vanilla CrazyFlie, include Flow deck, LED-ring deck, or Locopositioning deck, which can all be mounted down-facing on the CrazyFlie drone. Flow decks can enhance hovering and flight stability, using optical flow and Time-of-Flight (ToF) laser-ranging module. LED-ring deck is an array of 12 RGB LEDs that can be used to create light shows when grouped with other drones. Finally, the Locoposition deck can be used as part of an indoor localization system, which sends short high frequency radio messages between the Anchors and Tag (mounted on the Locoposition deck on the drone) to measure, through the means of triangulation, the absolute position of the drone.

For our work, Flow decks was mounted to the base our drones to ensure the stability of flight. This is because the current drivers that accompany the drones require this deck to maintain flight path and/or position.

%===============================================================================

\section{Detection Framework}
\label{sec:framework}

\begin{figure}
\begin{minipage}[b]{0.5\textwidth}
\centering
\includegraphics[width=0.9\textwidth, height=16cm]{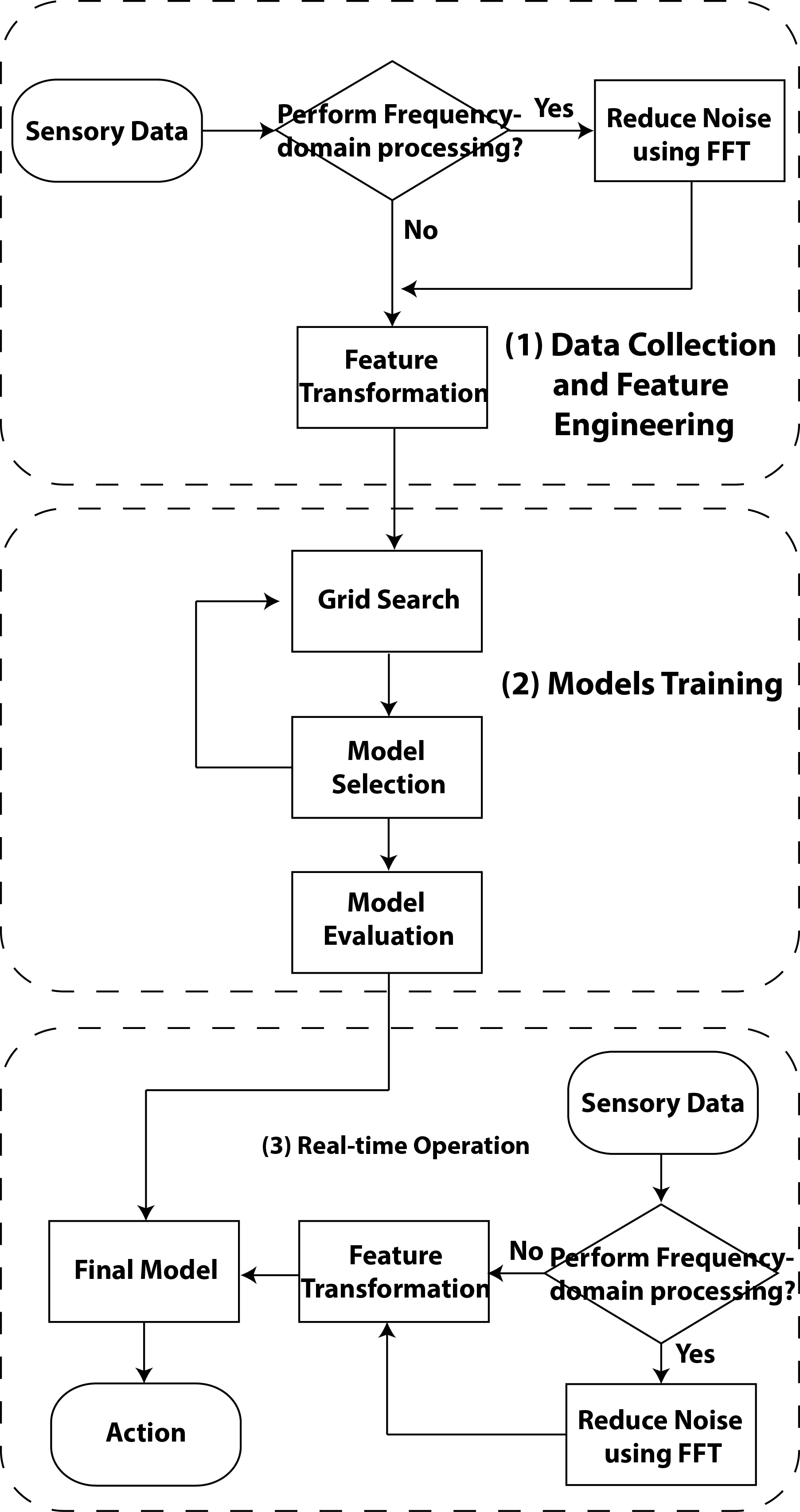}
\caption{Wind Gust Detection Framework}
\label{fig:flowchart}
\end{minipage}
\end{figure}

The physical sensor data as described in Section~\ref{sec:physical-sensors} are used to keep the drone stable and provide further information on its physical states. By utilizing such information for the purpose of environment sensing, we propose to push the limit of such data to achieve real-time wind gust detection. Fig~\ref{fig:flowchart} depicts the basic components of such a framework. 

Our framework first trains detection models during an offline phase and then applies such models in the real-time phase. In both offline training and real-time phases, further data pre-processing using Fast Fourier Transform (FFT) is applied in the hopes of boosting model performance. 

\subsection{Building a Wind Gust Detection Model}

Our framework is composed of two phases, an offline training phase (split into wind speed and direction detection) and an online real-time tracking phase. In the offline training phase, an environment sensing model is built to associate the physical sensory data as input with corresponding environment factors that affect the drones' stability. For instance, the model can capture the relationship between wind direction and strength, from the gyroscope and accelerometer readings. This task is similar to the works in activity recognition~\cite{ravi2005activity}, in which the goal is to associate human activities with such physical data. While for the online phase, the data is actively collected and analyzed immediately with serialized models from the offline phase.

In general, classification and regression models can be employed to achieve this task, depending on the nature of the environmental factors to be recognized. For example, wind strengths/speed can be categorized or regressed from input data, while wind direction may be categorized into different discrete directions and classified accordingly.

Furthermore, in real time wind gust detection, we look to use Fast Fourier transform (FFT) to help improve detection accuracy. In our study, FFT acts as a filter, as it removes the low frequencies that are unimportant given the effects of wind gust. We also propose a generalized method for applying FFT to our type of data by picking the most dominant “k” frequency domains from a sample of data and establishing out cutoff point for FFT with these frequency domains.

%===============================================================================

\section{Empirical Study}

To validate the feasibility of environment sensing using drones' physical sensors, we conducted an empirical study, in which the goal is to predict wind strengths given sensory data collected from the accelerometer and gyroscope. In the first study, we look at the feasibility of wind gust detection in terms of wind speed, wind direction and finally realtime analysis with pre-trained models while the drone is hovering. Then in study 2, we move to further testing of wind direction for drones in-motion.

\subsection{Data Collection}
\label{sec:data-collection}

We used CrazyFlie 2.0 drones as our target drones to collect data from. BitCraze's CrazyFlie 2.0~\cite{BitcrazeAB2016} is a family of palm-sized micro-drones that was designed for lightweight and versatile maneuver.

The data used in this study were collected from two main sensors, the accelerometer and gyroscope. A third set of data was collected from \textit{stabilization} values the drones outputted. The stabilizer values are also generated by the CrazyFlie drones which is the fusion of raw gyroscope and accelerometer data using an open-source algorithm called AHRS. 
Readings from the magnetometer (measuring orientation) and barometer (measuring pressure) are not studied as they are irrelevant to the stability of the drone when subject to wind gusts.

For Study 1(Section~\ref{sec:study-1}), the entire data set is collected through multiple flights due to the limited capacity of the battery. Since we will transform the data using a 1-second overlapping window (as discussed later in Section~\ref{sec:feat-eng}), total flight time for each packet was set at 64 seconds and we discarded 2 seconds of data points at the beginning and end of the each packet to make sure that no takeoff and landing turbulence may affect our data readings. This yields 6000 data points each flight data packet. For each wind speed or direction, a total of 6 packets were collected. 4 drones were used for wind speed and 3 were used for wind direction. Giving us a total of almost a million usable data points. 

To ensure consistency in data from one reading to the next, we made sure the drone stayed within 10 cm of its takeoff mark and discarded readings when the drone’s flight deviated outside of this area. Total flight time for each packet was set at 64 seconds and we also discarded 2 seconds of data points at the beginning and end of the each packet to make sure that no takeoff and landing turbulence may affect our data readings.

Similarly for Study 2 (Section~\ref{sec:study-2}), data was collected over multiple flights, however this time it was not only due to battery constraints, but also length of flight zone. Again we will transform the data using a 1-second overlapping time-window, and total flight time is about half a minute. Generating just over 2000 data points per data packet for each flight. Also to make up for the reduced data point from each flight, 10 packets were collected for each set of label instead of 6. 3 drones were used to collect data, but 1 drone’s dataset was corrupted, thus we had to discard it. However we still had over 200,000 usable data points from the 2 drones for data analysis in this study.

\subsubsection{Setup}

\begin{figure}
\begin{minipage}[b]{0.5\textwidth}
\includegraphics[width=0.9\textwidth]{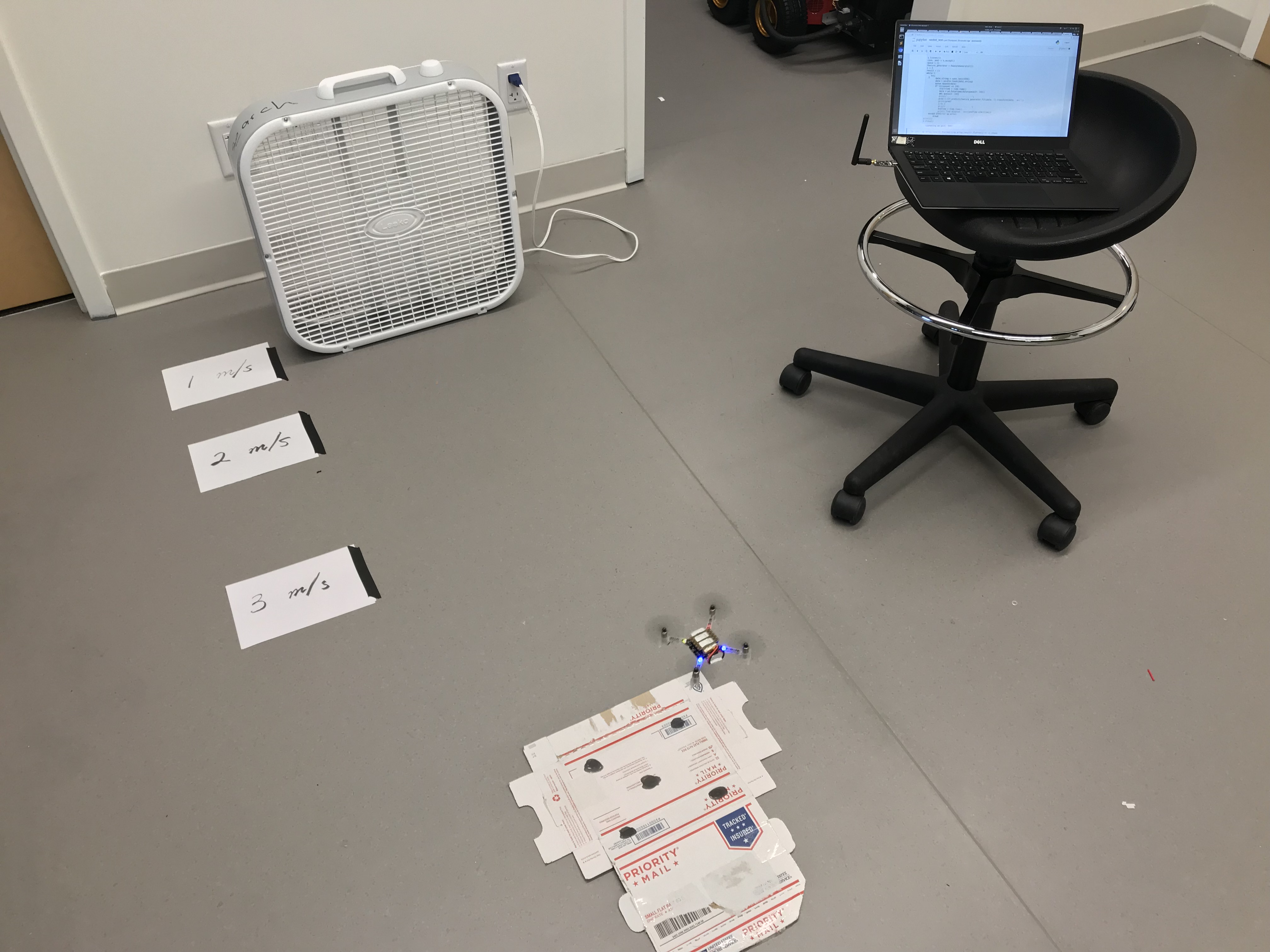}
\caption{Wind Gust Detection, Hovering}
\label{fig:setup1}
\end{minipage}
\end{figure}

To create air turbulence, we used a box fan, set up to blow air right into the drone. The setup for study 1 of the data collection is depicted in Fig~\ref{fig:setup1}. We collect sensory data from a hovering drone in four different conditions: No wind, Wind speed 1m/s (meters per second), Wind speed 2m/s and Wind speed 3m/s. The wind speeds are measured using a hand-held anemometer. The drone was also set to fly in a zone that has a consistent desired wind speed. For wind direction, the wind speed was set at 1.5 m/s, then the drone rotated in each direction to record head wind, tail wind, wind from left and right side. The head/front of the drone was determined by marking on the Crazyflie 2.0 that states the front of the drone.

Since we only wanted readings from the drone once it was hovering, we excluded data during take off and landing. The hovering height of the drones was set to 0.4 meters above the ground. This height value is selected such that the drone's hovering position lies right in the center of the air current produced by the fan. The drone was also set to fly at 0.5 meters away from the fan to ensure that the generated airflow covers a substantial area around the drone, and stays consistent throughout the data recording phase.

\begin{figure}
\begin{minipage}[b]{0.5\textwidth}
\includegraphics[width=0.9\textwidth]{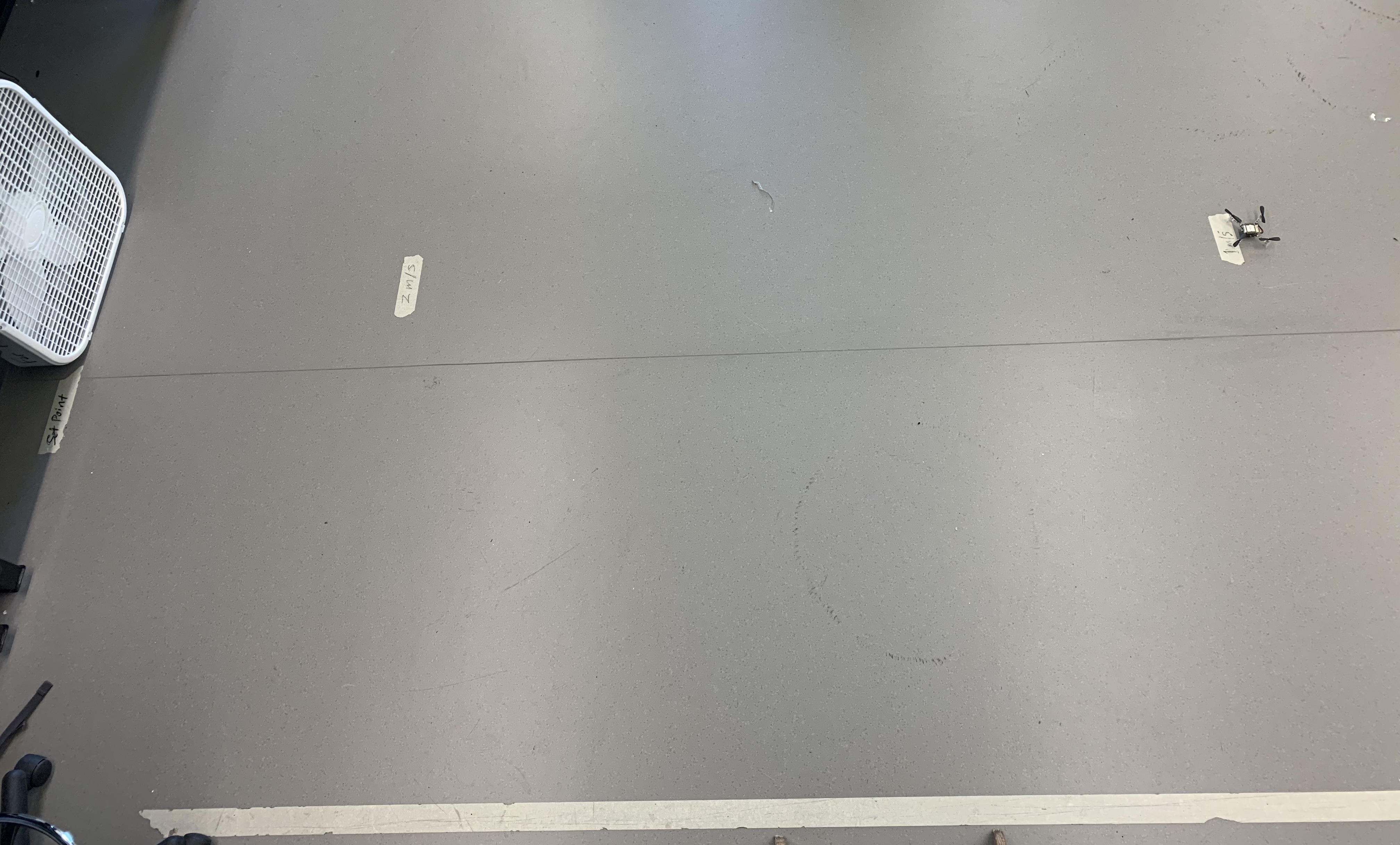}
\caption{Wind Direction Detection, In-motion}
\label{fig:setup2}
\end{minipage}
\end{figure}

For study 2, depicted in Fig~\ref{fig:setup2}. We collect sensory data from a drone in motion. It moves from a set point A (1m/s) to B (2m/s) marked on the floor, which we will call the flight zone. Thus for front/head wind, the drone will fly from point A to point B and back/tail wind, the drone will fly in the opposite direction (i.e point B to point A). For side wind, the box fan is set on the marked line at the bottom of the Fig~\ref{fig:setup2}, and it moves alone the with the drone as it flies from point A to B for left wind and vice versa for right wind.

This flight zone is limited between 1m/s and 2m/s wind speed when flying either against or away from the wind, because we wanted consistent data for each packet we collected. Thus if speed drops below 1m/s we feared the drone may not pickup the wind gust and if it goes above 2 m/s there is increase of turbulence which will throw the drone off course and render the packet of data useless. 

\subsubsection{Data Description}

The tables below show a snippet of the data we collected from the drones. The data are all time-stamped and for gyroscope (Table~\ref{tab:raw-data-gyro}) and accelerometer (Table~\ref{tab:raw-data-acc}) the three axes are x, y, and z. While for stabilizer (Table~\ref{tab:raw-data-stab}) the three axes are roll, pitch and yaw.

\begin{table}
\centering
\caption{Gyroscope (unit: degrees/s) - no wind data, label 0}
\label{tab:raw-data-gyro}
\begin{tabular}{|l|l|l|l|l|l|} 
\hline
\textbf{gyro.x}& \textbf{gyro.y}& \textbf{gyro.z} \\ 
\hline
-1.589356 & 12.612753 & -0.499988\\
\hline
0.026600 & 12.252775 & -0.453511\\ 
\hline
3.461869 & 2.456149 & -0.193813\\ 
\hline
4.814377 & -1.529192 & -0.630970\\ 
\hline
\end{tabular}
\end{table}

\begin{table}
\centering
\caption{Accelerometer (unit: g) - no wind data, label 0}
\label{tab:raw-data-acc}
\begin{tabular}{|l|l|l|l|l|l|} 
\hline
\textbf{acc.x}& \textbf{acc.y}& \textbf{acc.z} \\ 
\hline
-0.053416 & -0.027197 & 1.036873\\
\hline
-0.111874 & -0.074561 & 1.069046\\ 
\hline
-0.071886 & -0.065962 & 1.058163\\ 
\hline
0.000775 & -0.099285 & 0.980258\\ 
\hline
\end{tabular}
\end{table}

\begin{table}
\centering
\caption{Stabilizer (unit: degrees) - no wind data, label 0}
\label{tab:raw-data-stab}
\begin{tabular}{|l|l|l|l|l|l|} 
\hline
\textbf{stab.roll}& \textbf{stab.pitch}& \textbf{stab.yaw} \\ 
\hline
-0.188750 & 0.301978 & 0.158213\\
\hline
-0.176569 & 0.185277 & 0.152836\\ 
\hline
-0.141150 & -0.498941 & 0.151186\\ 
\hline
-0.069735 & -0.471636 & 0.146602\\ 
\hline
\end{tabular}
\end{table}

\begin{figure}
\begin{minipage}[b]{0.5\textwidth}
\includegraphics[width=1\textwidth]{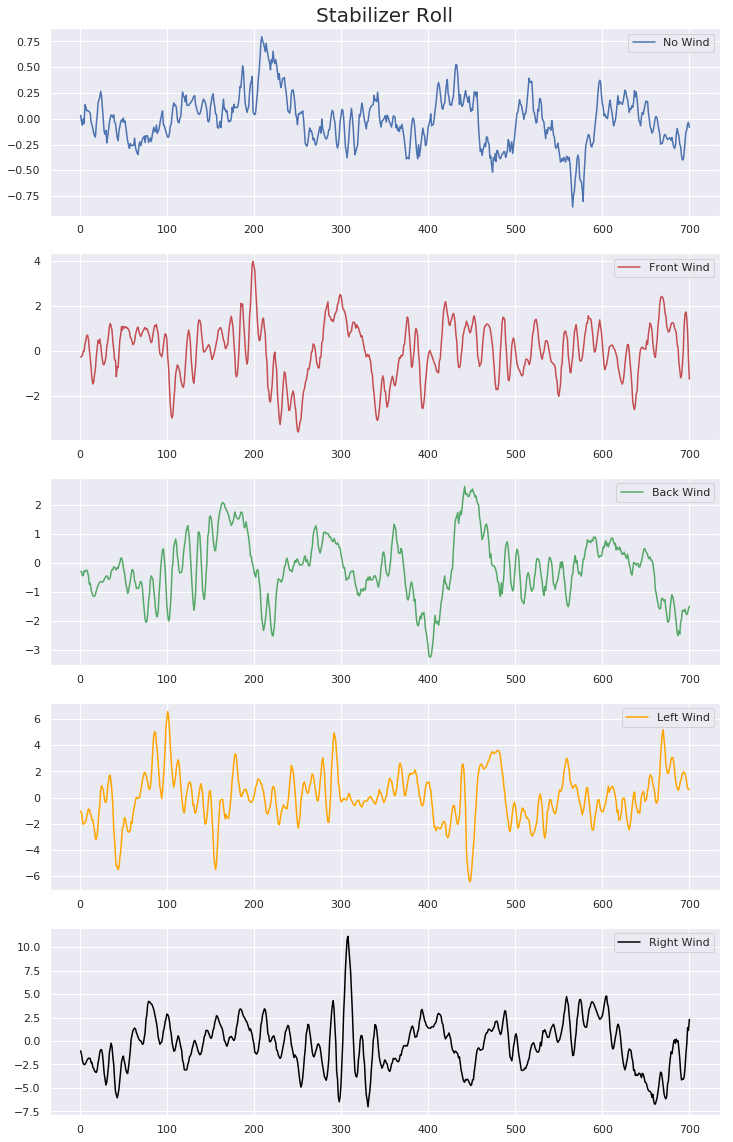}
\caption{Data Sample: Time-series Raw Data for Roll Column of Stabilizer}
\label{fig:data_sample1}
\end{minipage}
\end{figure}

\subsubsection{Feature Engineering}
\label{sec:feat-eng}

Since conventional machine learning models do not work well with time-series data, we pre-processed the data by segmenting it into examples using a sliding window of 1 second. Since the data is collected at 100Hz, one data window comprises of 100 raw data. Next, each data window from each sensor, i.e., accelerometer, gyroscope, and stabilizer, were used to extract 40 high-level features, including:
\begin{itemize}
    \item Mean (3): Mean sensor value (each axis).
    \item Standard deviation (3): Standard deviation (each axis).
    \item Average Absolute Difference (3): Average absolute difference between the 100 values and the mean of these values (each axis).
    \item Average Resultant Acceleration (1): For each of the 100 sensor values in the window, take the square root of x, y, and z axis values and then average them.
    \item Binned Distribution (30): 10 equal-sized bins are formed using the range of (maximum - minimum) of the 100 values, and record the fraction of the 100 values within each bin.
\end{itemize}
The reason for choosing such features is that, we look to preserve the trend of the data series with these simple hand crafted feature. Also in discussion with Dr. Weiss, we settle on these simple feature as more sophisticated mathematical features such as Fourier analysis did not yield much better results than these. In the end of feature transformation, we are able to keep an overall data trend while defining a more prominent averaging trend line.

\section{Study 1: Wind Gust Detection for Hovering Drone}
\label{sec:study-1}

In our first study, our goal is to examine the feasibility of detecting wind speeds and direction of the wind, i.e., the training and test data for a model come from the same drone, then the trained model is tested using test data from other drones to check generalization. Finally we look to test the real-time capabilities of our models.

\subsection{Model training}

We compare the performance of standard classification algorithms in detecting the wind condition under which observed data was recorded: Gradient Boosting (GB). We selected this models due to their superiority in modeling non-linear relationships in data ~\cite{weiss2016smartwatch}.

We split the entire data set into training test sets, with ratio 4:1, i.e., 80\% training and 20\% testing. We generate data features for the training and testing data separately. For each model, we applied grid search algorithm with 3-way hold out cross-validation to tune hyper-parameters on the training set. Finally, the left-out test set was used to evaluate each model. As for input features, we tested with data from accelerometer and gyroscope. In our study, we tested both classifier and regressor. For regressor, we measure model performance with r-squared value on a scale of 0 to 1, with any negative value representing an arbitrary bad model.

\subsection{Results}

\subsubsection{Wind Speed}
In wind speed detection, we achieved exceptional results using gyroscope data which can be seen in Table~\ref{tab:speed_result1}. Gyroscope Data using training and testing data from the same drone (within drone) yielded accuracies in the range of 85.9\% to 92.7\%. Even when using training data from one drone and testing with data from other drones (across drone), seen in Table~\ref{tab:speed_result2} we observe a high accuracy in the range of 80.2\% to 94.7\%. This proves the generalizability of the model for wind speed detection, in that we may use the same model across drones of the same class. In most cases when accuracy's are this high there isn't much to infer from the confusion matrix, however, if we observe the within drone and across drone confusion matrix Tables ~\ref{tab:speed_conf1} and ~\ref{tab:speed_conf2}, we notice a trend in the misclassifications. The number of misclassifications increase as our wind speed increases, this we believe, is due to the increase in turbulence at higher levels of wind speed, thus altering the data.

A reason why we did not test wind speed above 3 m/s is also attributed to the increase in turbulence at higher wind speed. At above 3 m/s, the drone would have a hard time taking off and maintaining it's flight position, thus making it impossible to collect meaningful consistent data for analysis. Also the reason for not breaking down the wind speeds into smaller increments is because in a previous experiment, we used much smaller increments of 0.4 m/s and noticed that there was quite a lot of misclassifications, especially across drone testing. This was because the change in data was not significant enough for the model to classify from one dataset to the other. 

\begin{table}
\centering
\caption{Study 1: Results Table 1.1, Accuracy within drones}
\label{tab:speed_result1}
\begin{tabular}{|l|l|l|l|l|l|} 
\hline
\textbf{Individual Drone} & 1 & 2 & 3 & 4 \\
\hline
\textbf{Gyroscope Accuracy (\%)} & 92.7 & 90.7 & 87.7 & 85.9 \\
\hline
\textbf{Accelerometer Accuracy (\%)} & 71.3 & 79.5 & 53.7 & 72.9 \\
\hline
\end{tabular}
\end{table}

\begin{table}
\centering
\caption{Study 1: Results Table 1.2, Gyroscope Accuracy 
across Drones}
\label{tab:speed_result2}
\begin{tabular}{|l|l|l|l|l|l|} 
\hline
\textbf{Train/Test} & Drone 1 & Drone 2 & Drone 3 & Drone 4 \\
\hline
\textbf{ Drone 1 (\%)} & - & 91.8 & 87.4 & 87.4 \\
\hline
\textbf{ Drone 2 (\%)} &  91.4 & - & 84.0 & 85.0 \\
\hline
\textbf{ Drone 3 (\%)} &  94.7 &  91.5 & - & 86.1 \\
\hline
\textbf{ Drone 4 (\%)} &  89.2 & 87.0 & 80.2 & - \\
\hline
\end{tabular}
\end{table}

\begin{table}
\centering
\caption{Confusion Matrix: Gradient Boosting, Wind Speed, Gyro, Drone 1 (Train/Test) Accuracy: 92.7\%}
\begin{tabular}{|l|l|l|l|l|l|} 
\hline
  & 0  m/s    & 1  m/s    & 2  m/s   & 3 m/s    \\ 
\hline
0  m/s & 7101 & 0    & 0    & 0  \\ 
\hline
1  m/s & 0    & 6814 & 287  & 0     \\ 
\hline
2  m/s & 0    & 193  & 6666 & 242     \\ 
\hline
3  m/s & 0    & 0  & 1362  & 5739   \\ 
\hline
\end{tabular}
\label{tab:speed_conf1}
\end{table}

\begin{table}
\centering
\caption{Confusion Matrix: Gradient Boosting, Wind Speed, Gyro, Drone 1 Train / Drone 3 Test Accuracy: 94.7\%}
\begin{tabular}{|l|l|l|l|l|l|} 
\hline
  & 0 m/s    & 1 m/s    & 2 m/s   & 3  m/s   \\ 
\hline
0  m/s & 7101 & 0    & 0    & 0  \\ 
\hline
1  m/s & 0    & 6756 & 345  & 0     \\ 
\hline
2  m/s & 0    & 211  & 6232 & 658     \\ 
\hline
3  m/s & 0    & 0  & 291  & 6810   \\ 
\hline
\end{tabular}
\label{tab:speed_conf2}
\end{table}

\subsubsection{Wind Direction}
In wind direction detection, the accelerometer accuracy was much higher, all above 95\%. However, gyroscope data left much to desired, being as low as 31.3\%, which can be seen in Table~\ref{tab:direc_result1}. However, when testing data across drone for accelerometer data, we see a large variation in accuracy ranging from as low ass 51.8\% to 80.5\%. Thus we would like to look at the confusion matrix to see if we can find a trend with why the classifier performed poorly in across drone analysis. However when looking at the confusion matrices for the across drone analysis, Table~\ref{tab:direc_conf1} show most of the misclassifications in back wind classified as front wind, while in Table~\ref{tab:direc_conf2} the misclassifications are in completely different and random areas.

Since the drones in this study are hovering, there should be no resultant acceleration. Thus accelerometer data should have had no influence on the outcome. Furthermore, the high accuracy of accelerometer data for within drone classification can be attributed to over-fitting of the classifier to the minute changes in data when adjusting to wind from different direction and also shown in the high training data accuracy, while having much lower test accuracy when testing with different data.

This led us to move forward in our data collection, by collecting directional data for drones in-motion which we dive into in study 2.

\begin{table}
\centering
\caption{Study 1: Results Table 2.1, Accuracy within drones}
\label{tab:direc_result1}
\begin{tabular}{|l|l|l|l|l|l|} 
\hline
\textbf{Individual Drone} & 1 & 2 & 3 \\
\hline
\textbf{Gyroscope Accuracy (\%)} &  45.1 & 54.7 & 31.3  \\
\hline
\textbf{Accelerometer Accuracy (\%)} &  95.8 & 98.2 & 95.6  \\
\hline
\end{tabular}
\end{table}

\begin{table}
\centering
\caption{Study 1: Results Table 2.2, Accelerometer Accuracy 
across Drones}
\label{tab:direc_result2}
\begin{tabular}{|l|l|l|l|l|l|} 
\hline
\textbf{Train/Test} & 1 & 2 & 3 \\
\hline
\textbf{ 1 (\%)} & - & 78.4 & 80.5 \\
\hline
\textbf{ 2 (\%)} &  51.8 & - & 65.6 \\
\hline
\textbf{ 3 (\%)} &   70.2 & 69.6 & - \\
\hline
\end{tabular}
\end{table}

\begin{table}
\centering
\caption{Confusion Matrix: Gradient Boosting, Wind Direction, Accelerometer, Drone 1 Train / Drone 3 Test Accuracy: 70.2\%}
\begin{tabular}{|l|l|l|l|l|l|} 
\hline
  & front    & back    & left    & right \\
\hline
front & 6679 & 4370    & 0    & 943  \\ 
\hline
back & 0    & 388 & 186  & 53     \\ 
\hline
left & 71    & 2272  & 6862 & 72     \\ 
\hline
right & 422    & 69  & 0  & 6017   \\ 
\hline
\end{tabular}
\label{tab:direc_conf1}
\end{table}

\begin{table}
\centering
\caption{Confusion Matrix: Gradient Boosting, Wind Direction, Accelerometer, Drone 3 Train / Drone 2 Test Accuracy: 65.6\%}
\begin{tabular}{|l|l|l|l|l|l|} 
\hline
  & front    & back    & left    & right \\
\hline
front & 3768 & 0    & 1209    & 0  \\ 
\hline
back & 0    & 6123 & 3343  & 0     \\ 
\hline
left & 0    & 0  & 1643 & 0     \\ 
\hline
right & 3332    & 977  & 905  & 7100   \\ 
\hline
\end{tabular}
\label{tab:direc_conf2}
\end{table}

\subsubsection{Real-time Analysis}
For real-time analysis we built a system that collects the data from the drones and outputs a prediction in real time. For our experimental purposes, the setup itself is similar to that of the offline phase. We then marked the different wind speeds on the ground shown in Fig~\ref{fig:setup1} and changed the location of the fan accordingly. Once the drone is hovering, we collect data in small batches of a 100 data points (rows). We then apply FFT and feature generation on the data. Finally, we test the data using saved serialized models from the offline training phase and record the outputs over a one-minute interval. This is done to allow us to draw accuracy reading from the testing. However, to fully test the real-time capability of our models, we vary the wind speed actively and seeing how the model responded. A video footage of our real-time detection experiment is provided on YouTube\footnote{\url{https://youtu.be/hZRcSMr0DFE}}.

We were able to achieve an output of result at an average rate of one every 0.04 seconds (i.e. 25Hz). Real-time testing proved the strength of our wind speed detection model, where we were able to achieve 100\% accuracy when no wind was present, 92\%, 86\% and 98\% accuracy for wind speeds 1m/s, 2m/s and 3 m/s respectively. 

However, the accuracy of wind direction detection in real time was no better than guessing when testing. The randomness in prediction when varying the direction of wind acting on the drone meant we weren't able to collect any meaningful data for analysis.

\section{Study 2: Wind Direction Detection for In-motion Drone}
\label{sec:study-2}
In our second study, our goal is to examine the feasibility of detecting wind direction of the drones when in-motion. This study aims to extend the analysis from study 1, specifically the directional analysis model.

\subsection{Model Testing}
We employ similar method of testing the data from study 1. But with stabilizer data. This is due to observations made in study 1, were hovering drone meant that there was no resultant acceleration. Thus with the introduction of motion into the analysis, we must include both gyroscope and accelerometer data, and since Stabilizer data is a fuse of the 2 as mentioned above, it naturally became the best data type for this study.

\subsection{Results}
For the 2 drones used in study 2, we received an accuracy of 62.36\% and 57.81\% for drone 1 and 2 respectively. This is much lower than the accuracy we desired, which should be around 90\% to allow accurate response of the drones to the predictions. However, when looking at confusion matrices in Tables~\ref{tab:direc_conf3} and ~\ref{tab:direc_conf4} for both drones, we notice the diagonal line of correct classification still contains the majority of the data points, but there are quite a lot of misclassifications throughout and concentrated in the left and right wind area. So the question becomes, why the model failed for this data, but also what was the model classifying. We will break it down to 2 methodologies of explainging this. 1, Data Similarity and 2, Power Spectral Density.

\begin{table}
\centering
\caption{Confusion Matrix: Study 2, Gradient Boosting, Wind Direction, Stabilizer, Drone 1 Overall Accuracy: 62.36\%}
\begin{tabular}{|l|l|l|l|l|l|} 
\hline
  & No wind & front    & back    & left    & right \\
\hline
No wind & 3143 & 168    & 378    & 22 & 58  \\ 
\hline
front & 240 & 2311    & 505    & 341 & 364  \\ 
\hline
back & 36    & 602 & 1553  & 932 & 633     \\ 
\hline
left & 9    & 520  & 146 & 2078 & 1003     \\ 
\hline
right & 76    & 91  & 149  & 799 & 2666   \\ 
\hline
\end{tabular}
\label{tab:direc_conf3}
\end{table}

\begin{table}
\centering
\caption{Confusion Matrix: Study 2, Gradient Boosting, Wind Direction, Stabilizer, Drone 2 Overall Accuracy: 57.81\%}
\begin{tabular}{|l|l|l|l|l|l|} 
\hline
  & No wind & front    & back    & left    & right \\
\hline
No wind & 3281 & 297    & 189    & 2 & 0  \\ 
\hline
front & 8 & 2463    & 958    & 226 & 125  \\ 
\hline
back & 76    & 612 & 2005  & 676 & 409     \\ 
\hline
left & 186    & 212  & 645 & 1982 & 753     \\ 
\hline
right & 1    & 895  & 504  & 1194 & 1187   \\ 
\hline
\end{tabular}
\label{tab:direc_conf4}
\end{table}

\subsubsection{Data Similarity}
Thus to test mathematically if there is any opposing trend between the different directions of wind (i.e. left vs right, front vs back) to explore 3 method of calculating the similarity in data. These being Euclidean Distance, Dynamic Time Warping and Cross Correlation. For Euclidean Distance and Dynamic Time Warping because there is no meaningful benchmark for which we can compare the opposing datasets, we thus decide to use Cross Correlation. With Cross Correlation, correlation at lag of 0 (i.e. X lags mean we shift a sequence to the left or right by x displacement.), data is normalized by the correlation model and resulting values fall between -1 and 1 for easy interpretation. Were -1 is negative similarity and 1 is strong positive similarity.

However when viewing the resulting correlation scores for the different directions in Table~\ref{tab:correlation}, we notice that all score hover around 0, which means that there is no similarity or correlation between the different directional data, thus there was models were not able to find a opposing trend between the directions.

\begin{table}
\centering
\caption{Correlation Scores: Study 2, Cross Correlation}
\begin{tabular}{|l|l|l|l|l|l|} 
\hline
  & No wind & front    & back    & left    & right \\
\hline
No wind & 1 & 0.0892    & 0.0151    & -0.0099 & 0.0977  \\ 
\hline
front & - & 1    & 0.0426    & 0.0255 & 0.0397  \\ 
\hline
back & -    & - & 1  & -0.0298 & -0.0319   \\ 
\hline
left & -    & -  & - & 1 & -0.0822     \\ 
\hline
right & -    & -  & -  & - & 1   \\ 
\hline
\end{tabular}
\label{tab:correlation}
\end{table}

\subsubsection{Power Spectral Density and Fourier Transform}
Finally we get to the Power Spectral Density/Frequency Domain analysis to figure out what exactly was the models classifying. Power Spectral Density (PSD) is a frequency-domain plot of power/Hz vs frequency. Power Density tells us which frequencies contain the signal’s power, measure in amplitude\^2/Hz. We employed Fourier Transform to decompose the data into sinusoidal components. Then, we used the components to further calculate the amplitude at each frequency.

\begin{figure}
\begin{minipage}[b]{0.5\textwidth}
\includegraphics[width=1\textwidth]{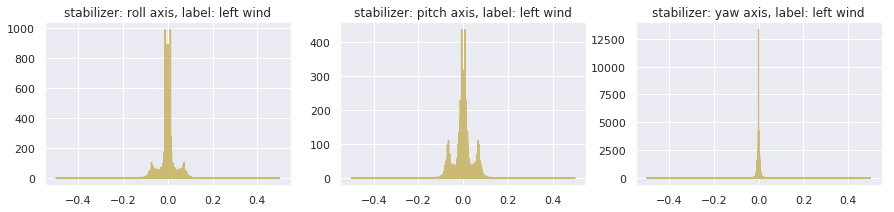}
\includegraphics[width=1\textwidth]{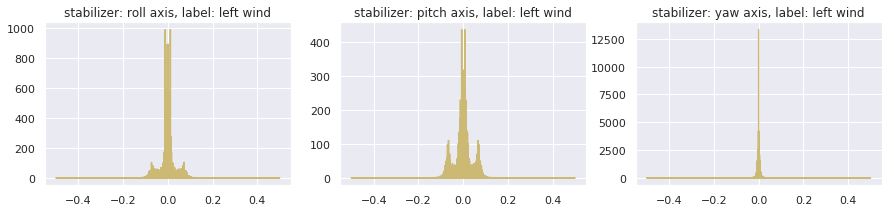}
\includegraphics[width=1\textwidth]{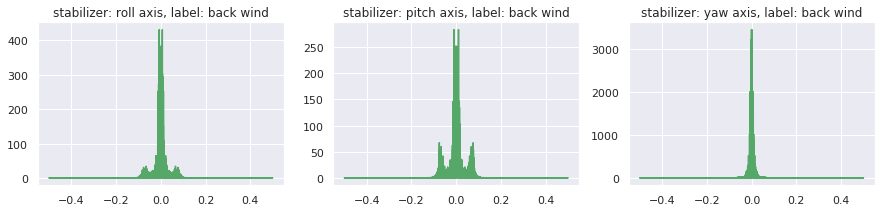}
\includegraphics[width=1\textwidth]{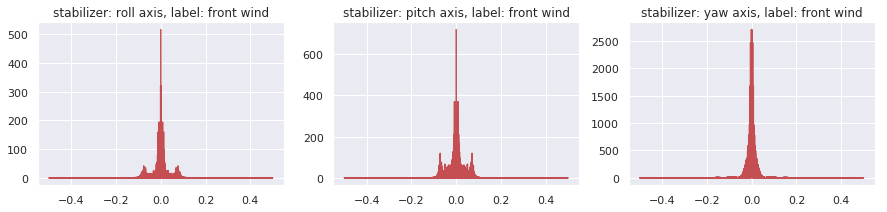}
\includegraphics[width=1\textwidth]{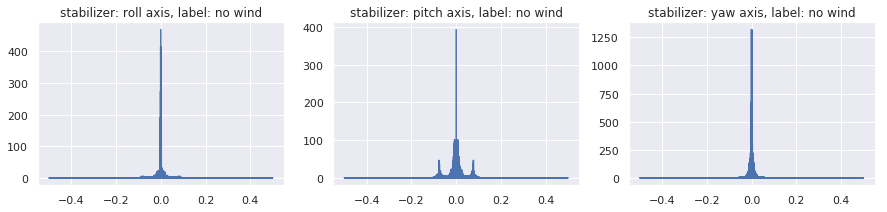}
\caption{Frequency Domain of Different Directional Wind Stabilizer Data }
\label{fig:freq_domain}
\end{minipage}
\end{figure}

If we look at the frequency domain visualization of the different direction of wind in Fig~\ref{fig:freq_domain}. We see that for no wind, front and back wind the amplitude of the power spectrum is around 400 to 500 for roll data. While for left and right wind the amplitude is much higher. This explains the miss classifications, as if we ignore the higher amplitude of data, were only small sets of data points lie, we notice that there is a large number of data points that lie within the same amount of change from 0 power/Hz (the equilibrium point) to 500 power/Hz for all directions. Thus we can conclude for this study, the classification models were classifying the amplitude of the sinusoidal components in data and because of this, with many data point for all directions lying in the same deviation from equilibrium point, the classifier will miss classify the data points.

\section{Supplemental Analysis and Testing}
For study 1 we also applied Fast Fourier Transformation  ~\ref{subsec:fft} to our data to test if it will increase our accuracy and in study 2, we used a Convolutional Neural Network ~\ref{subsec:cnn}to test our raw dataset. 

\subsection{Fast Fourier Transform (FFT)}
\label{subsec:fft}
FFT acts as a filtering mechanism after analyzing the frequency domain of the dataset. We determine our cutoff point for FFT in the frequency domain by taking the k most dominant number of data points from the 100 data point time window, ~\ref{tab:fft_table2}. The Table~\ref{tab:fft_table1} shows the results of applying FFT to our raw dataset of Drone 1, with marginal increase in accuracy, especially at the cost of increase in computational intensity. Furthermore, when the dataset became large enough, there was no increase in accuracy observed when applying FFT. However, for smaller dataset like when dealing with real-time analysis, when we are only able to analyze 100 data point at once, FFT could prove valuable.

\begin{table}
\caption{Drone 1, Accuracy Summary,  Wind Speed, FFT k Selection}
\begin{tabular}{|c|c|c|c|c|c|}
    \hline
    \multicolumn{6}{|c|}{Drone1, Random Forest Classifier} \\ \hline
    k         & 0      & 5      & 10     & 20     & 30     \\ \hline
    accuracy  & 0.9171 & 0.9828 & 0.9089 & 0.9129 & 0.9185 \\ \hline
\end{tabular}
\label{tab:fft_table2}
\end{table}

\begin{table}
\caption{Drone 1, Accuracy Summary, Wind Speed Detection (FFT)}
\begin{tabular}{c|c|c|}
    \cline{2-3}
                                                & \multicolumn{2}{c|}{Accuracy}             \\ \cline{2-3} \centering
                                                       & No FFT Applied    & FFT Applied, k = 30   \\ \hline
    \multicolumn{1}{|c|}{RF Classifier}     & 0.9171            & 0.9185                \\ \hline
    \multicolumn{1}{|c|}{RF Regressor}      & 0.9504            & 0.9494                \\ \hline
    \multicolumn{1}{|c|}{GB Classifier} & 0.9260            & 0.9273                \\ \hline
    \multicolumn{1}{|c|}{GB Regressor}  & 0.9525            & 0.9515                \\ \hline
\end{tabular}
\label{tab:fft_table1}
\end{table}

\subsection{Neural Network, Convolutional Neural Network (CNN) with Tensorflow}
\label{subsec:cnn}
\begin{figure}
\begin{minipage}[b]{0.5\textwidth}
\centering
\includegraphics[width=0.9\textwidth]{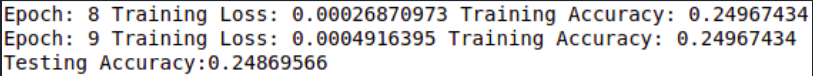}
\caption{CNN Testing Results}
\label{fig:cnn_fig}
\end{minipage}
\end{figure}

In Study 2, we explored the use of Convolutional Neural Network (CNN) as it allows the model itself to choose the most optimal features from raw data. The CNN model consists of one convolution layer followed by max pooling and another convolution layer. After that, the model will have fully connected layer which is connected to Softmax layer. The model was based our model off of human activity recognition cnn models \footnote{\url{http://aqibsaeed.github.io/2016-11-04-human-activity-recognition-cnn/}}. However the results from our initial analysis with CNN left much to be desired, which can be seen in the Fig~\ref{fig:cnn_fig} showing an accuracy of just 24.87\%.

\section{Conclusion}
In this paper, we propose a new methodology for wind detection in quadcopters (drones). Our proposed approach has demonstrated its instrumentality in within-drone and across-drone wind speed detection, making it possible to apply such detection in real-time. We believe this framework will be useful for MAVs/UAVs when operating in areas with high air turbulence, helping them avoid fatal crashes due to strong wind gusts, without the need to add additional bulky sensors. This will also help reduce weight thus increasing battery life of drone flight which is very much limited at this stage of the technology. 
This project has set a great benchmark for further work in the future in this area of testing. Especially if we can alter the features for direction detection by futher research into Deep Neural Networks and better raw dataset. It is to our knownledge that the Fordham Robotics Lab recently implemented ROS(Robot Operating System) drivers for Crazyflie 2.0 drones, which we had no access to before. This will allow us to collect more specific data and manipulate the drone in a more advanced ways.

\section{Thank You}
Finally we would like to thank Dr. Li Yanjun for all the help and guidance during this capstone project course, raising valuable questions about our methodologies, so as to help us find the right answers to our problems.
We would also like to thank  Dr. Lyons, Dr. Nguyen and the Fordham Robotics Lab for their support, equipment and guidance throughout this project.

%%%%%%%%%%%%%%%%%%%%%%%%%%%%%%%%%%%%%%%%%%%%%%%%%%%%%%%%%%%%%%%%%%%%%%%%%%%%%%%%

\bibliographystyle{IEEEtran}
\bibliography{wind_detection_references} 

\end{document}